# XLS-R Deep Learning Model for Multilingual ASR on Low- Resource Languages: Indonesian, Javanese, and Sundanese

Panji Arisaputra[1,*], Alif Tri Handoyo[1] and Amalia Zahra[2]

[1]Computer Science Department, School of Computer Science
Bina Nusantara University
Jakarta 11480, Indonesia
alif.handoyo@binus.ac.id; *Corresponding author: panji.arisaputra@binus.ac.id

[2] Computer Science Department, BINUS Graduate Program, Master of Computer Science
Bina Nusantara University
Jakarta, 11480, Indonesia
amalia.zahra@binus.edu

ABSTRACT. *This research paper focuses on the development and evaluation of Automatic Speech Recognition (ASR) technology using the XLS-R 300m model. The study aims to improve ASR performance in converting spoken language into written text, specifically for Indonesian, Javanese, and Sundanese languages. The paper discusses the testing procedures, datasets used, and methodology employed in training and evaluating the ASR systems. The results show that the XLS-R 300m model achieves competitive Word Error Rate (WER) measurements, with a slight compromise in performance for Javanese and Sundanese languages. The integration of a 5-gram KenLM language model significantly reduces WER and enhances ASR accuracy. The research contributes to the advancement of ASR technology by addressing linguistic diversity and improving performance across various languages. The findings provide insights into optimizing ASR accuracy and applicability for diverse linguistic contexts.*

1. **Introduction.** ASR is a technological innovation that automatically converts verbal translations into written texts. It focuses on reducing Word Error Rate (WER) metrics when reproducing oral input. ASR's core capability is to act as an optimal connector for information exchange between human-to-human and human-to-machine entities [1]. It has become increasingly important in various domains, including air traffic control, biometric security, games, closed text for YouTube, voice message transcription, and home automation. ASR's implementation in digital media resources is not a new phenomenon, but its complexity has increased [2].

This study focuses on the rapid development of information and communication technology in Indonesia. In Figure 1, the data from the Central Statistics Agency (*Badan Pusat Statistik* (BPS)) [3] shows that 62.10% and 82.07% of Indonesians have access to the internet in 2021, followed by an increase in mobile phone use of 65.87%. However, less mobile technology is being abandoned, such as computers and cable phones, which are only 18.24% and 1.36%, respectively. The conclusion is that Indonesians are shifting from traditional technology to more mobile and agile devices like smartphones, which require the right modalities for effective and efficient operation.

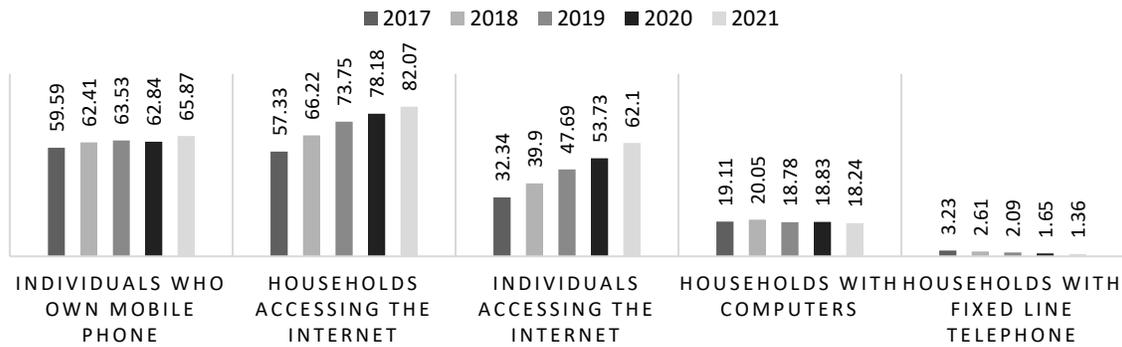

FIGURE 1. Development of information and communication technology
In Indonesia 2017-2021 (value in percent(%))

Yu and Deng [1] argue that speech recognition plays a crucial role in facilitating the advancement of more flexible technology. Hence, the development of ASR is imperative, as speech represents the most innate form of communication among individuals, surpassing written or typed interactions. In contrast, several factors exert a significant influence on the precision and efficacy of ASR systems. These factors include accent, speaker class characteristics, and the utilization of dialect, all of which have been shown to impact ASR performance [4].

This paper will primarily examine the testing procedures involved in developing ASR systems for languages such as Indonesian, Javanese, and Sundanese, in addition to the work conducted by Arisaputra and Zahra [5]. According to Kjartansson, Sarin, Pipatsrisawat, Jansche, and Ha [6], Javanese and Sundanese are identified as the second and third most widely spoken languages in Indonesia, following Indonesian, boasting an estimated 90 and 40 million speakers, respectively.

Arisaputra and Zahra [5] employed the pre-trained XLSR-53 model [7] in their study. However, in this paper, we intend to evaluate alternative deep learning-based approaches by fine-tuning the XLS-R model [8], incorporating its Transformer architecture, on various datasets, including TITML-IDN [9], Magic Data (Indonesian Scripted Speech Corpus—Daily Use Sentence), Common Voice [10], OpenSLR—Large Javanese & Sundanese ASR training data set (SLR35 & SLR36) [6], and OpenSLR—High-quality TTS data for Javanese & Sundanese (SLR41 & SLR44) [11].

Maxwell-Smith, Z., & Foley, B. [12] carried out research on Indonesian ASR using the most recent pre-trained model by using XLSR-53 [7] and IndonesianNLP [13] (a fine-tuned version of the XLSR-53 model [7] on the Indonesian Common Voice dataset) for training with the Online Indonesian Learning (OIL) dataset [14]. Maxwell-Smith, Z., & Foley, B. [12] concluded that the results clearly show that existing pre-trained Indonesian models are not suitable for processing language teaching data holistically. Also, Maxwell-Smith, Z., & Foley, B.'s [12] investigation into the potential advantages of employing a multilingual language model for enhancing outcomes, as said by San et al. [15], was hindered by the intricate nature of the dataset's languages. Additional research is required to advance the development of intricate multilingual language models that are compatible with the aforementioned languages. Furthermore, further investigation is warranted to assess the effectiveness of employing complex language model in ASR systems. Hence, it is of great scientific interest and presents a significant challenge to explore potential solutions for overcoming this obstacle.

The wav2vec 2.0 model [16] is a comprehensive approach to developing robust speech representations using unsupervised methods. It uses a Convolutional Neural Network (CNN) [17] as its fundamental feature encoder and demonstrates superior performance compared to previous models. The XLSR-53 model, on the other hand, employs self-supervised learning techniques to generate extensive cross-lingual speech representations, using product quantization and Gumbel-Softmax to choose discrete codebook entries. This model is similar to BERT's [18] contextual network but with a different positional embedding method.

The lack of datasets has slowed the advancement of ASR, particularly for the Indonesian language. Recent research by Prakoso, Ferdiana, & Hartanto [19] showed notable advancements in speech recognition performance with their experiment using the CMUSphinx Toolkit and a custom language model for training. Their training yielded a WER of 14%, while testing resulted in a 20% WER within a 27,764 dB environment.

The XLS-R framework, proposed by Babu et al. [8], builds upon the principles of wav2vec 2.0 and has an impressive scale of 2 billion parameters derived from an extensive dataset of nearly half a million hours of audio data across 128 languages. This model has shown significant reductions in WER, ranging from 14% to 34%. The effectiveness of cross-lingual pretraining can be comparable to that of pretraining exclusively in English, given that the model size is adequate.

The XLS-R model was chosen for this study to improve ASR technology's performance. The XLS-R model with 300 million parameters, referred to as XLS-R 300m, will be used. The study aims to reduce WER measurements and enhance ASR outcomes in converting spoken language into written text, focusing on Indonesian, Javanese, and Sundanese languages. The XLS-R model is used on various datasets to train and evaluate ASR systems, improving their robustness. A comparative assessment is conducted to evaluate and compare the proposed XLS-R 300m model with existing models, including the XLSR-53 model used by Arisaputra and Zahra [5].

2. **Methodology.** This section is dedicated to outlining the data employed for the study. A comprehensive elucidation of the research's procedural trajectory, commencing from its inception and culminating in its conclusion, will also be comprehensively furnished within this section.

2.1. **Dataset.** As previously stated, a total of seven datasets were utilized in this study. Three of these datasets were specific to the Indonesian language, namely TITML-IDN, Magic Data (Indonesian Scripted Speech Corpus—Daily Use Sentence), and Common Voice. The remaining four datasets were focused on Javanese and Sundanese languages, including OpenSLR—Large Javanese & Sundanese ASR training data set (SLR35 & SLR36), as well as OpenSLR—High-quality TTS data for Javanese & Sundanese (SLR41 & SLR44).

**TITML-IDN.** The dataset consists of speech data from 20 Indonesian speakers, resulting in a phonetically balanced collection of 343 sentences. The methodology used an established text corpus for text-based Information Retrieval (IR), with additional sentences added to enhance the corpus. The Information and Language Processing System (ILPS) group curates the foundational corpus, which is openly accessible to the public. Sentences were selected and organized according to specific criteria, including 1,000 sentences with various recognition units, 300 sentences with less frequently occurring units, and 43 sentences with higher-frequency elements. The cohort included 20 individuals from Indonesia's five most prominent ethnic groups. Audio recordings were obtained in a controlled acoustic setting, resulting in a speech corpus lasting 14.5 hours [9].

**Magic Data (Indonesian Scripted Speech Corpus—Daily Use Sentence).** Beijing Magic Data Technology Company Limited is an international provider of AI data services, including speech recognition, speech synthesis, and Natural Language Processing (NLP) research. They offer an open-source dataset called Magic Data, which comes from the MagicHub platform. The details can be seen in Table 1, which contains a transcription of 3.5 hours of Indonesian scripted speeches from 10 people. The data operates at a sampling rate of 16 kHz, eliminating the need for data resampling and making it a valuable resource for AI developers.

TABLE 1. The details about the Magic Data

| Speaker ID | Gender | Age | Region | Device |
| --- | --- | --- | --- | --- |
| G0004 | M | 26 | Jakarta | HONORRVL-AL09 |
| G0005 | M | 25 | Jakarta | XiaomiRedmi Note 4 |
| G0006 | M | 28 | Jakarta | XiaomiMi A1 |
| G0007 | M | 29 | Jakarta | XiaomiRedmi Note 5 |
| G0008 | M | 21 | East Java | HUAWEIBLA-AL00 |
| G0111 | F | 20 | Sulawesi | SamsungSM-J730G |
| G0112 | F | 25 | Jakarta | OPPOA37f |
| G0113 | F | 21 | Jakarta | OPPOA37f |
| G0114 | F | 23 | Bali | MotorolaXT1663 |
| G0115 | F | 23 | Jakarta | SamsungSM-J710F |

**Common Voice.** Mozilla's Common Voice project, which includes over 2 million hours of audio recordings in 38 languages, aims to create a global repository for speech recognition. The project involves volunteers recording and evaluating sentences, and the corpus is licensed under the CC0 public domain. The dataset was acquired from Hugging Face and divided into six subsets: train, validation/dev, test, other, validated, and invalidated. This study used three subsets: train, validation/dev, and test, with 5,809 instances from 170 individuals. The data set is divided into two categories: males (62%), and females (38%). The age distribution of the population is as follows: individuals under 19 years old make up 39%, those aged 19 to 29 make up 43%, those aged 30 to 39 make up 11%, those aged 40 to 49 make up 5%, and those aged 50 to 59 make up 2%. The dataset is subject to a standardized sampling rate of 48 kHz, necessitating resampling to achieve a 16 kHz sampling rate.

**OpenSLR—Large Javanese & Sundanese ASR training data set (SLR35 & SLR36).** As can be seen in Table 2, this corpus consists of about 200,000 speech recordings from native speakers in two regions who volunteered for the study. Data was collected using consumer electronic devices in a controlled environment, with speakers representing a variety of gender identities. Each corpus contains audio files and a .tsv file named utt_spk_text.tsv, which are arranged in a structured manner.

TABLE 2. The details about the OpenSLR—Large Javanese & Sundanese ASR training data set (SLR35 & SLR36)

| ID | .zip File Name | Status | Language | Recordings | Hours | Speakers |
|---|---|---|---|---|---|---|
| SLR35 | asr_javanese_0…2 | Used | Javanese | 185,076 | 296 | 1,019 |
| | asr_javanese_3…9 | Not Used | | | | |
| | asr_javanese_1…f | Not Used | | | | |
| SLR36 | asr_sundanese_0…2 | Used | Sundanese | 219,156 | 333 | 542 |
| | asr_sundanese_3…9 | Not Used | | | | |
| | asr_sundanese_1…f | Not Used | | | | |

The recordings, a direct result of community volunteers' efforts, highlight the immeasurable value of their contributions. The Javanese and Sundanese datasets have 16 .zip files available, but each was limited to three due to computational capacity constraints. The audio was recorded using a 16-bit linear Pulse Code Modulation format and stored in the Free Lossless Audio Codec (FLAC) format to preserve quality. The datasets are denoted as SLR35 and SLR36 [6].

**OpenSLR—High quality TTS data for Javanese & Sundanese (SLR41 & SLR44).** In Table 3, detailed information about the unique characteristics and storage details of the audio corpus pertaining to each language has been given, the dataset consists of pairs of audio recordings and corresponding transcripts representing a unique language. Each pair is matched with a standardized textual transcription, resulting in an audio content representation in the RIFF WAVE format. The audio data compilation involved volunteers aged 20–22 representing various linguistic contexts. Participants were required to articulate succinct sentences from reputable sources. Recording sessions were conducted in acoustically controlled settings with stringent Quality Control (QC) protocols to preserve audio fidelity, reduce ambient noise, and synchronize audio with text transcripts. The datasets will be denoted as SLR41 and SLR44 [11].

TABLE 3. The details about the OpenSLR—High quality TTS data for Javanese & Sundanese (SLR41 & SLR44)

| ID | Languages | Gender | Speaker Count | Number of Audio Files | Total Duration (Hours) |
|---|---|---|---|---|---|
| SLR41 | Javanese | Male | 21 | 2958 | 3:28:13 |
| | | Female | 20 | 2864 | 3:31:13 |
| SLR44 | Sundanese | Male | 22 | 1812 | 2:10:22 |
| | | Female | 21 | 2401 | 3:12:21 |

2.2. **Deep Learning with XLS-R 300m.** This experiment is divided into 4 parts: (1) pre-processing data; (2) fine-tuning; (3) building and embedding the language model; and (4) evaluation, as illustrated in Figure 2.

**Pre-processing data.** The datasets TITML-IDN, Magic Data, Common Voice, SLR35, SLR36, SLR41, and SLR44 will be divided into 90% train data and 10% test data. The train data will be divided into two subsets: a train set and a val set. These subsets will be combined into unified divisions and matched with their corresponding transcriptions. The audio content will be standardized into the .wav format, adhering to specifications like a single channel configuration and a 16 kHz sampling rate. A normalization procedure will remove special characters from transcriptions, retaining only the lowercase alphabet. A vocabulary will be created by extracting unique letters from the train and val set data. A mapping function will be developed to integrate all transcriptions into a cohesive extended transcription. The parameter 'batched = True' will be used to ensure comprehensive access to all transcriptions simultaneously. The training and validation data will undergo consolidation and transformation, resulting in an enumerated dictionary containing a wide range of letters.

To improve clarity, the " " character, representing a blank token, will be replaced with a visually distinguishable character like "|". An "unknown" token will enhance the model's ability to handle unfamiliar characters. Token padding will be implemented using the Connectionist Temporal Classification (CTC) algorithm, which accommodates the "blank token" The tokenizer will be systematically established using the Wav2Vec2CTCTokenizer class from the Hugging Face Transformers library after quantifying tokens in the data [20].

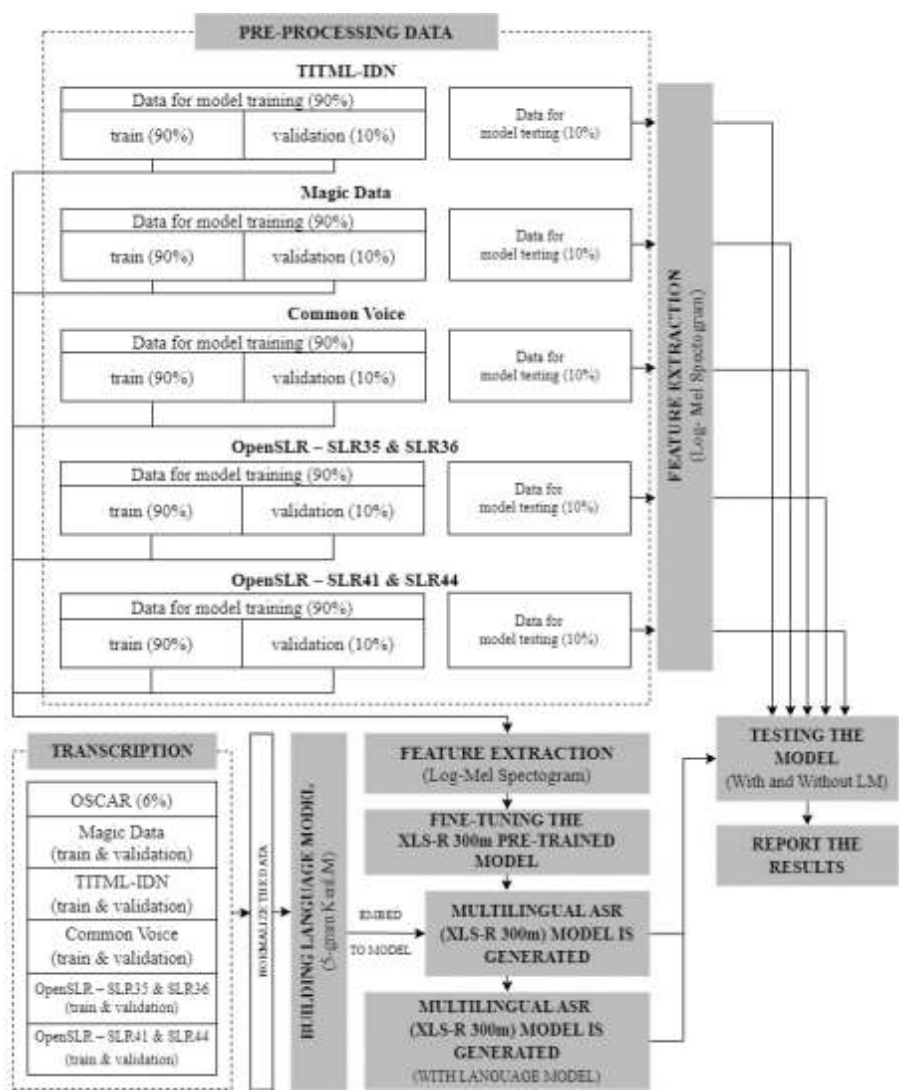

FIGURE 2. Proposed Methodology

**Fine-tuning**. The process of building a processor involves combining the tokenizer and feature extractor components. The tokenizer is created after data preparation, and the focus then shifts to the feature extractor. The feature encoder converts features into more compact and informative representation vectors, reducing the need for complex model processing. Before moving on to the fine-tuning stage, which the CTC algorithm facilitates, the preliminary stage is crucial.

Data sampling is essential before exploring the feature extractor, and the sampling rate determines the frequency of voice signal data points captured per second. The Wav2Vec2FeatureExtractor class from the Hugging Face Transformers library is invoked to instantiate the feature extractor. The tokenizer and feature extractor are combined into a single entity called the Wav2Vec2Processor class, which serves as the model's specialized processor.

Fine-tuning involves integrating a supplementary classifier on top of the model, representing the output vocabulary for each task. The classifier is trained using the CTC loss technique [20, 21], and the weight of the encoder feature remains unchanged throughout the fine-tuning stage. The learning rate used in this study is 3e–4, with a batch size of 5 and a total of 10 epochs. The evaluation process is performed at 200 steps, with the gradient_accumulation_steps parameter set to 2.

**Building and embedding the language model**. The study uses KenLM [22] as a language model, incorporating text corpora from four sources: OSCAR (using 6%), TITML-IDN, Magic Data, and Common Voice. OSCAR is used as an augmentation technique due to its large size, which can significantly influence the language model's results. The subset "unshuffled_deduplicated_id" contains 2,394,957,629 Indonesian words, with only 6% of the transcription data used for this study. This approach acknowledges the importance of the large corpus in influencing the language model's results.

**Evaluation**. For the evaluation, we will compare KenLM-enhanced models with non-enhanced models using the WER metric on a test dataset. The trials were conducted by generating multiple KenLM variants, specifically 2-gram, 3-gram, 4-gram, 5-gram, and 6-gram models. Additionally, it will use the XLSR-53 model to compare the current model to a previous model and calculate the mean WER produced by each model in order to determine which model is the most effective. This comparative analysis will help identify the most effective model.

3. **Results.** Table 4 shows that the XLS-R 300m model did about as well as the XLSR-53 model in an experiment, with a WER of 10.38% on the Magic Data dataset and just a little bit better than the XLSR-53 model. In the TITML-IDN dataset, XLSR-53 emerged as the dominant model, exhibiting a slightly superior WER of 0.72%. However, the XLS-R 300m maintains its competitive edge by supporting two additional languages, Javanese and Sundanese, albeit with a slight compromise in performance.

Even though both the XLSR-53 and XLS-R 300m models use the same Common Voice dataset, there are no comparable results for the XLSR-53 model on the Common Voice dataset because of differences in the methods used to split up the data. Hence, the assessment of WER outcomes on the Common Voice dataset is considered inconsequential. In the case of XLSR-53, three out of the six subsets, specifically the train, validation/dev, and test subsets, are utilized in their original state without undergoing any form of amalgamation. In the Common Voice dataset, the default allocation of data is as follows: 36.37% is designated for the train, 31.59% is allocated for the validation/dev, and 31.74% is reserved for the test. The data volume affects how Arisaputra and Zahra [5] allocated the data for their study. The train set comprises 2,130 instances, while the validation/dev set comprises 1,835 instances.

Additionally, the test set consists of 1,844 instances. On the other hand, this study using the XLS-R 300m model merged the Common Voice data set and then divided it into three parts: train, val, and test, as illustrated in Figure 2. The assessment of XLSR-53's performance on Javanese and Sundanese languages is also considered irrelevant, as XLSR-53 is specifically designed for Indonesian and not intended to be used for the aforementioned languages like Javanese and Sundanese.

According to the data presented in Table 4, the XLS-R 300m model demonstrates the most favorable WER of 5.43% on average. This achievement is attributed to the incorporation of a 5-gram KenLM into the fine-tuned model.

The integration of a language model results in a significant reduction in the WER, indicating a substantial advancement. One crucial factor in the development of language models is the size of the transcription, as a larger transcription leads to a language model of higher quality. In addition, the length of the context in which a language model operates has an impact on its quality. Integrating a higher 'n' in n-grams into the ASR model leads to enhanced model performance. Nevertheless, it should be noted that an abundance of contexts does not necessarily ensure improved performance [23].

TABLE 4. Comparison results for each model.
Metric use WER (%) and Language Model use KenLM

| Model | Data Train & Val | KenLM | Data Testing | | | | | | | AVG |
|---|---|---|---|---|---|---|---|---|---|---|
| | | | TITML-IDN | Magic Data | Common Voice | SLR 35 | SLR 36 | SLR 41 | SLR 44 | |
| XLS-R 300m ASR Multi-lingual Model | • TITML-IDN<br>• Magic Data<br>• Common Voice<br>• SLR35<br>• SLR36<br>• SLR41<br>• SLR44 | - | 7.73 | 19.64 | 15.30 | 17.95 | 2.39 | 21.99 | 7.10 | 13.16 |
| | | 2-gram | 1.79 | 10.93 | 6.55 | 7.76 | 1.20 | 10.90 | 3.58 | 6.10 |
| | | 3-gram | 1.39 | **10.38** | 5.63 | 6.50 | 1.15 | 10.41 | 3.47 | 5.56 |
| | | 4-gram | 1.37 | **10.38** | 5.11 | **6.38** | 1.15 | 10.31 | 3.47 | 5.45 |
| | | 5-gram | 1.37 | **10.38** | 4.99 | 6.41 | **1.14** | **10.25** | **3.44** | **5.43** |
| | | 6-gram | 1.37 | **10.38** | 5.01 | 6.41 | **1.14** | 10.35 | **3.44** | 5.44 |
| XLSR-53 ASR Model [5] | • TITML-IDN<br>• Magic Data<br>• Common Voice | - | 2.17 | 16.75 | - | - | - | - | - | 9.46 |
| | | 2-gram | 0.77 | 10.78 | - | - | - | - | - | 5.77 |
| | | 3-gram | **0.72** | 10.88 | - | - | - | - | - | 5.80 |
| | | 4-gram | **0.72** | 10.88 | - | - | - | - | - | 5.80 |
| | | 5-gram | **0.72** | 10.93 | - | - | - | - | - | 5.82 |

4. **Conclusions.** This research investigates ASR technology with an emphasis on optimizing accuracy and performance with the pre-trained XLS-R 300m model. Using the XLSR-53 model, Arisaputra and Zahra [5] designed an ASR system customized for Indonesian society. The applicability of the model to languages other than Indonesian, such as Javanese and Sundanese, remained unaddressed. ASR research has been preoccupied with comparative evaluations of ASR models, particularly on diverse data sets.

The study made novel contributions to the advancement of ASR technology by evaluating the performance of the XLS-R 300m model on various datasets containing Indonesian, Javanese, and Sundanese languages. The incorporation of a 5-gram KenLM into the optimized model led to a significant decrease in the WER. Furthermore, the research recognized the importance of the quality of language models and sought to enhance performance by increasing the value of 'n' in n-grams, although this approach did not consistently yield improvements.

In conclusion, this study builds upon Arisaputra and Zahra's [5] work and significantly extends our understanding of ASR technology's capabilities. It acknowledges linguistic diversity and leverages advanced techniques like 5-gram KenLM integration. By delving into language models, dataset partitioning, and contextual length, the research presents a robust framework for improving ASR accuracy across various languages, enhancing its applicability and performance in diverse linguistic contexts.

Further studies will be conducted with the objective of enhancing ASR performance, with a specific focus on low-resource languages. These investigations will center around the examination of the effects of various data augmentation techniques, including but not limited to Pitch Shift, Velocity Perturbation, Noise Injection, Echo Simulation, Language Interpolation, Transliteration, and Voice Cloning, on ASR performance. In addition, further studies will expand the evaluation by testing and comparing pre-trained models with a wider spectrum. This extension seeks to further advance our understanding of the adaptability and accuracy of ASR technology in diverse language contexts, ultimately contributing to the evolution of ASR systems for a broader range of languages and applications.